# DeepPrior++: Improving Fast and Accurate 3D Hand Pose Estimation

Markus Oberweger[1]     Vincent Lepetit[1,2]

[1]Institute for Computer Graphics and Vision, Graz University of Technology, Austria
[2]Laboratoire Bordelais de Recherche en Informatique, Université de Bordeaux, France

{oberweger,lepetit}@icg.tugraz.at

## Abstract

*DeepPrior [18] is a simple approach based on Deep Learning that predicts the joint 3D locations of a hand given a depth map. Since its publication early 2015, it has been outperformed by several impressive works. Here we show that with simple improvements: adding ResNet layers, data augmentation, and better initial hand localization, we achieve better or similar performance than more sophisticated recent methods on the three main benchmarks (NYU, ICVL, MSRA) while keeping the simplicity of the original method. Our new implementation is available at https://github.com/moberweger/deep-prior-pp.*

## 1. Introduction

Accurate hand pose estimation is an important requirement for many Human Computer Interaction or Augmented Reality tasks, and has attracted lots of attention in the Computer Vision research community [9, 10, 16, 20, 21, 34, 36, 43]. Even with 3D sensors such as structured-light or time-of-flight sensors, it is still very challenging, as the hand has many degrees of freedom, and exhibits self-similarity and self-occlusions in images.

One popular method for 3D hand pose estimation is *DeepPrior*, introduced by [18]. DeepPrior is a Deep Network-based approach that uses a single depth image as input and directly predicts the 3D joint locations of the hand skeleton. The key idea in DeepPrior is to explicitly integrate a prior on 3D hand poses computed by Principal Component Analysis (PCA) directly into a Convolutional Neural Network. This offers a simple, yet accurate and fast method for 3D hand pose estimation.

Since the publication of the original paper, there has been tremendous advances in the field of Machine Learning and Deep Neural Networks. We leverage recent progress in this field and update the original approach. Thus we call the resulting approach *DeepPrior++*. Specifically:

- we updated the model architecture to make the model more powerful by introducing a Residual Network [7] for extracting feature maps;

- we improved the initial hand localization method. This step in DeepPrior was based on a heuristics. Here we use a trained method;

- we improved the training procedure to leverage more information from the available data.

We released the code with our improvements at https://github.com/moberweger/deep-prior-pp with the hope that it will be useful for the community.

In the following, we shortly review the original DeepPrior approach in Section 3, then introduce our modifications in Section 4. The modifications are evaluated in Section 5 with a comparison to state-of-the-art methods on public benchmark datasets.

## 2. Related Work

There is a significant amount of early work that deals with hand pose estimation, and we refer to [3] for an overview. In 2015, an evaluation of several works on benchmark datasets [30] has shown that DeepPrior performed state-of-the-art in terms of accuracy and speed. There have been tremendous advances since then, and here we shortly review related works. We compare against all the works that report results using the commonly used error metrics on at least one of the three major benchmark datasets, *i.e.* NYU [40], MSRA [28], and ICVL [34]. These works are marked in this section with a star *.

Many recent approaches exploit the hierarchy of the hand kinematic tree. [35]* proceeds along the skeleton tree and predicts the positions of the child joints within the tree. Similarly, [44] (Lie-X)* predicts updates along the skeleton tree that correct an initial pose and use Lie-algebra to constrain these updates. Sun *et al*. [28] (HPR)* estimate the joint locations in normalized coordinate frames for each finger, and [25] uses a separate regressor for each finger to predict spatial and temporal features that are combined in a nearest-neighbor formulation. [46] introduces a spatial attention mechanism that specializes on each joint and an additional optimization step to enforce kinematic constraints. [14] splits the hand into smaller sub-regions along the kinematic tree. [45]* predicts a gesture class for each pose and trains a separate pose regressor for each class. All these approaches require multiple predictors, one for each joint or finger, and often additional regressors for different iterations of the algorithms. Thus the number of regression models ranges from tens to more than 50 different models that have to be trained and evaluated.

To overcome this shortcoming, there are several works that integrate the kinematic hierarchy into a single CNN structure. Guo *et al*. [6] (REN)* train an ensemble of sub-networks for different spatial regions of input features, and Madadi *et al*. [15]* use a tree-shaped CNN architecture that predicts different parts of the kinematic tree. However, this requires a specifically designed CNN architecture depending on the annotation.

Different data representations of the input depth image were also proposed. Deng *et al*. [2] (Hand3D)* convert the depth image to a 3D volume and use a 3D CNN to predict joint locations. However, 3D networks show a low computational efficiency [23]. Differently, [42]* uses surface normals instead of the depth image, but surface normals are not readily accessible from current depth sensors and thus introduce an additional computational overhead. Neverova *et al*. [17]* combine a segmentation of the hand parts with a regression of joint locations, but the segmentation is sensitive to the sensor noise.

Instead of predicting the 3D joint locations directly, [40]* proposed an approach to predict 2D heatmaps for the different joints. [5]* extended this work and use multiple CNNs to predict heatmaps from different reprojections of the depth image, which requires a separate CNN for each reprojection. Also, these approaches require complex post-processing to fit a kinematic model to the heatmaps.

A probabilistic framework was proposed by Bouchacourt *et al*. [1] (DISCO)*, who use a network to learn the posterior distribution of hand poses and one can sample from this distribution. However, it is unclear how to combine these samples in practice. Wan *et al*. [41] (Crossing Nets)* use two generative networks, one for the hand pose and one for the depth image, and learn a shared mapping between these two networks, which involves training several networks in a complex procedure.

Oberweger *et al*. [19] (Feedback)* learn a CNN to synthesize depth image of a hand and use the synthesized depth image to predict updates for an initial hand pose. Again, this requires training three different networks.

Zhou *et al*. [48] (DeepModel)* integrate a hand model into a CNN, by introducing an additional layer that enforces the physical constraints of a 3D hand model, where the constraints have to be manually defined beforehand.

Fourure *et al*. [4] (JTSC)* exploit different annotations from different datasets by introducing a shared representation, which is an interesting idea for harvesting more training samples, but has shortcomings when dealing with sensor characteristics.

Zhang *et al*. [47]* formulate pose estimation as a multivariate regression problem that, however, requires solving a complex optimization problem during runtime.

There are also generative model-based approaches that recently raised much attention. Although being very accurate, the works of [12]*, [26], [32]*, [37]* require a 3D model of the hand, which should be adjusted to the users' hand [26], [33]*, and run a complex optimization during inference.

Comparing to these recent approaches, our method is easier and faster to train, has a simpler architecture, is more accurate, and runs at a comparable speed, *i.e*. realtime.

## 3. Original DeepPrior

In this section, we briefly review the original DeepPrior method. More details can be found in [18].

DeepPrior aims at estimating the 3D hand joint locations from a single depth image. It requires a set of depth images labeled with the 3D joint locations for training.

To simplify the regression task, DeepPrior first performs a 3D detection of the hand. It then estimates a coarse 3D bounding box containing the hand. Following [34], DeepPrior assumes the hand is the closest object to the camera, and extracts a fixed-size cube centered on the center of mass of this object from the depth map. It then resizes the extracted cube to a $128 \times 128$ patch of depth values normalized to $[-1, 1]$.

Points for which the depth is not available—which may happen with structured light sensors for example—or the depth values are farther than the back face of the cube, are assigned a depth of 1. This normalization is important for the learning stage in order to be invariant to different distances from the hand to the camera.

Given the physical constraints over the hand, there are strong correlation between the different 3D joint locations. Instead of directly predicting the 3D joint locations, DeepPrior therefore predicts the parameters of the pose in a lower dimensional space. As this enforces constraints of the hand pose, this improves the reliability of the predictions.

As shown in Figure 1, DeepPrior implements the pose prior into the network structure by initializing the weights of the last layer with the major components from a PCA of the 3D hand pose data. Then, the full network is trained using standard back-propagation.

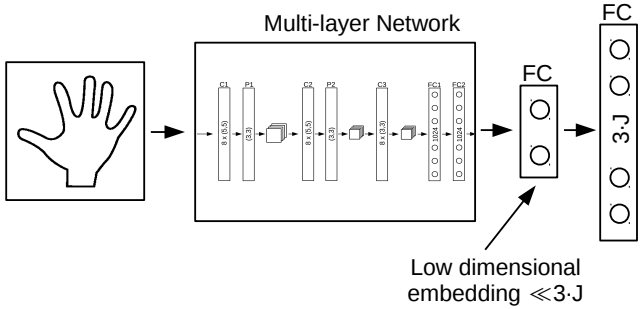

Figure 1: The network architecture for the original DeepPrior. C denotes a convolutional layer with the number of filters and the filter size inscribed, FC a fully-connected layer with the number of neurons, and P a max-pooling layer with the pooling size. The shown Multi-layer Network can be an arbitrary Neural Network, with an additional layer for the prior. DeepPrior introduces a pose prior by precomputing the weights of the last layer from a PCA applied to the 3D hand pose data.

## 4. DeepPrior++

In this section, we describe our changes to enhance the original DeepPrior approach, which includes improved training data augmentation, better hand localization, and a more powerful network architecture. For implementation level details we refer to the code.

### 4.1. Improved Training Data Augmentation

Since our approach is data-driven, we aim at leveraging as much information as possible from the available data. There have been many different augmentation methods used in literature [13, 40], such as scaling, flipping, mirroring, rotating, *etc*. In this work we use depth images, which give rise to specific data augmentation methods. Specifically, we use rotation, scaling, and translation, as well as different combinations of them.

**Rotation:** The hand can be rotated easily around the forearm. This rotation can be approximated by simple in-plane rotation of the depth images. We use random in-plane rotations of the image, and change the 3D annotations accordingly by projecting the 3D annotations onto the 2D image, applying the same in-plane rotation, and projecting the 2D annotations back to 3D coordinates. The rotation angle is sampled from a uniform distribution with the interval $[-180°, 180°]$.

**Scaling:** The MSRA [28] and NYU [40] datasets contain different persons, with different hand size and shape. Although DeepPrior is not explicitly invariant to scale, we can train the network to be invariant to hand size by varying the size of the crop in the training data. Therefore, we scale the 3D bounding box for the crop from the depth image by a random factor sampled from a normal distribution with mean of 1 and variance of $0.02$. This changes the appearance of the hand size in the cropped cube, and we scale the 3D joint locations according to the random factor.

**Translation:** Since the hand 3D localization is not perfect, we augment the training set by adding random 3D offsets to the hand 3D location, and center the crops from the depth images on these 3D locations. We sample the random offsets from a normal distribution with a variance of 5mm, which is comparable to the error of the hand 3D detector we use. We also modify the 3D annotations according to this offset.

**Online Augmentation:** The augmentation is performed online during training and thus the network sees different samples at each epoch. This leads to more than 10M different samples in total. The augmentation helps to prevent overfitting and to be more robust to deviations of the hand from the training set. Although the samples are correlated, it significantly helps at test time, as we show in the experiments.

**Robust Prior:** Similarly, we also improve the prior, which is obtained by applying PCA to the 3D hand poses. We sample 1M poses, by randomly using rotation, scaling, and translation of the original poses in 3D. We use this augmented set of 3D poses for calculating the prior.

### 4.2. Refined Hand Localization

The original DeepPrior used a very simple hand detection. It was based on the center of mass of the depth segmentation of the hand. Therefore, the hand was segmented using depth-thresholding, and the 3D center of mass was calculated. Then, a 3D bounding box was extracted around the center of mass.

DeepPrior++ still uses this method but introduces a refinement step that significantly improves the final accuracy. This refinement step relies on a regression CNN. This CNN is applied to the 3D bounding box centered on the center of mass, and is trained to predict the location of the Metacarpophalangeal (MCP) joint of the middle finger, which we use as referential. We also use augmented training data to train this CNN as described in Section 4.1.

For real-time applications, instead of extracting the center of mass from each frame, we apply this regression CNN to the hand location of the previous frame. This remains accurate while being faster in practice.

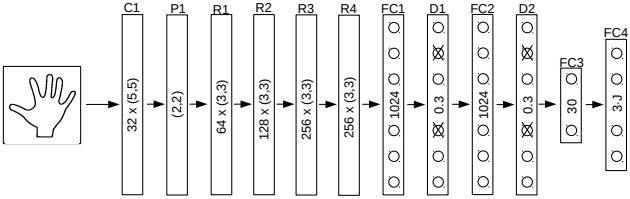

Figure 2: Our ResNet architecture. C denotes a convolutional layer with the number of filters and the filter size inscribed, FC a fully-connected layer with the number of neurons, D a Dropout layer with the probability of dropping a neuron, R a residual module with the number of filters and filter size, and P a max-pooling layer with the pooling region size. The hand crop from the depth image is fed to the ResNet that predicts the final 3D hand pose.

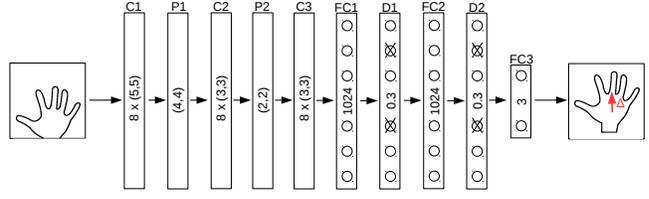

Figure 3: The network architecture used for refining hand localization. As in Fig. 2, C denotes a convolutional layer, FC a fully-connected layer, D a Dropout layer, and P a max-pooling layer. The initial hand crop from the depth image is fed to the network that predicts an offset to correct an inaccurate hand localization.

### 4.3. More Powerful Network Architecture

**Residual Networks.** Since the introduction of DeepPrior, there has been much research on better deep architectures [8, 24, 31], and the Residual Network (ResNet) architecture [8] appears to be one of the best performing models.

Our model is similar to the 50-layer ResNet model of [8]. Since ResNet was originally proposed for image classification, we adapt the architecture to fit our regression problem. Most importantly, we remove the global average pooling, and add two fully-connected layers. The input to the network is $128 \times 128$ pixel, with values normalized to $[-1, 1]$. The adapted ResNet model is shown in Figure 2. The network contains an initial convolution layer with 64 filters and $2 \times 2$ max-pooling. This convolutional layer is followed by four residual modules, each with a stride of $2 \times 2$, and with $\{64, 128, 256, 256\}$ filters.

The much simpler model used for refining the hand localization is shown in Fig. 3. It consists of three convolutional layers with max-pooling, and two fully-connected layers with Dropout.

We optimize the network parameters using the gradient descent algorithm ADAM [11] with standard hyperparameters and a learning rate of $0.0001$, and train for 100 epochs.

**Regularization using Dropout.** The ResNet model can overfit, and we experienced this behavior especially on datasets with small hand pose variation [34]. Therefore, we introduce Dropout [27] to the model, which was shown to provide an effective way of regularizing a neural network. We apply binary Dropout with a dropout rate of $0.3$ on both fully-connected layers after the residual modules. This enables training high capacity ResNet models while avoiding overfitting and achieving highly accurate predictions.

## 5. Evaluation

We evaluate our DeepPrior++ approach on three public benchmark datasets for hand pose estimation: the NYU dataset [40], the ICVL dataset [34], and the MSRA dataset [28]. For the comparison with other methods, we focus here on works that were published after the original DeepPrior paper. There are different evaluation metrics used in the literature for hand pose estimation, and we report the numbers stated in the papers or measured from the graphs if provided, and/or plot the relevant graphs for comparison.

For all experiments, we report the results for a 30-dimensional PCA prior. By using an efficient implementation for data augmentation, the training time is the same for all experiments, approximately 10 hours on a computer with an Intel i7 with 3.2GHz and 64GB of RAM, and an nVidia GTX 980 Ti graphics card.

### 5.1. Evaluation Metrics

We use two different metrics to evaluate the accuracy:

- First, we evaluate the accuracy of the 3D hand pose estimation as average 3D joint error. This is established as the most commonly used metric in literature, and allows comparison with many other works due to simplicity of evaluation.

- As a second, more challenging metric, we plot the fraction of frames where all predicted joints are below a given maximum Euclidean distance from the ground truth [38].

### 5.2. NYU Dataset

The NYU dataset [40] contains over 72k training and 8k test frames of multi-view RGB-D data. The dataset was captured using a structured light-based sensor. Thus, the depth maps show missing values as well as noisy outlines, which makes the dataset very challenging. For our experiments we use only the depth data from a single camera. The dataset has accurate annotations and exhibits a high variability of different poses. The training set contains samples from a single user and the test set samples from two different users. We follow the established evaluation protocol [18, 40] and use the 14 joints for calculating the metrics.

Our results are shown in Table 1 together with a comparison to current state-of-the-art methods. We compare DeepPrior++ to several related methods, and it significantly outperforms the other methods.

| Method | Average 3D error |
| --- | --- |
| Oberweger *et al.* [18] (DeepPrior) | 19.8mm |
| Oberweger *et al.* [19] (Feedback) | 16.2mm |
| Deng *et al.* [2] (Hand3D) | 17.6mm |
| Guo *et al.* [6] (REN) | 13.4mm |
| Bouchacourt *et al.* [1] (DISCO) | 20.7mm |
| Zhou *et al.* [48] (DeepModel) | 16.9mm |
| Xu *et al.* [44] (Lie-X) | 14.5mm |
| Neverova *et al.* [17] | 14.9mm |
| Wan *et al.* [41] (Crossing Nets) | 15.5mm |
| Fourure *et al.* [4] (JTSC) | 16.8mm |
| Zhang *et al.* [47] | 18.3mm |
| Madadi *et al.* [15] | 15.6mm |
| **This work (DeepPrior++)** | **12.3mm** |

Table 1: Comparison with state-of-the-art on the NYU dataset [40]. We report the average 3D error in mm. DeepPrior++ significantly performs better than all other methods for this dataset.

In Figure 4 we compare our method with other discriminative approaches. Although Supancic *et al.* [30] report a very accurate results for a fraction of the frames, our approach significantly performs better for the majority of the frames.

In Figure 5 we compare state-of-the-art methods using a different evaluation protocol, *i.e.* we follow the protocol of [32, 37], who evaluate the first 2400 frames of the test set. Also for this protocol, we significantly outperform the state-of-the-art method of Taylor *et al.* [37]. Note that [32, 33, 37] require a, possibly user-specific, 3D hand model, whereas our method only uses training data without any 3D model.

### 5.3. ICVL Dataset

The ICVL dataset [34] comprises a training set of over 180k depth frames showing various hand poses. The test set contains two sequences with each approximately 700 frames. The dataset is recorded using a time-of-flight camera and has 16 annotated joints. The depth images have a high quality with hardly any missing depth values,

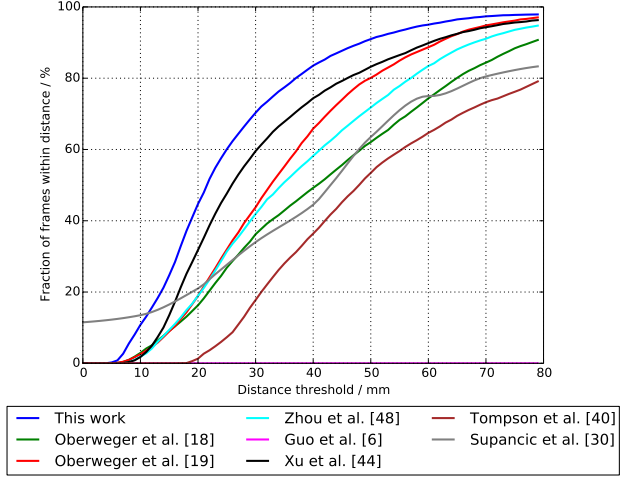

Figure 4: Comparison with state-of-the-art discriminative methods on the NYU dataset [40]. We plot the fraction of frames where all joints are within a maximum distance from the ground truth. A larger area under the curve indicates better results. Our proposed approach performs best among other discriminative methods. (Best viewed in color)

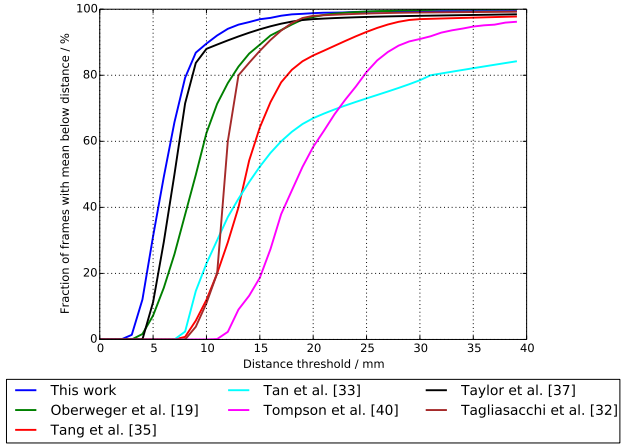

Figure 5: Comparison with state-of-the-art model-based methods on the NYU dataset [40]. We plot the fraction of frames where the average joint error per frame is within a maximum distance from the ground truth, following the protocol of [32, 37]. A larger area under the curve indicates better results. Our proposed approach even outperforms model-based approaches on this dataset, with more than 90% of the frames with an error smaller than 10mm. (Best viewed in color)

and sharp outlines with little noise. Although the authors provide different artificially rotated training samples, we start from the genuine 22k frames only and apply the data augmentation as described in Section 4.1. However, the pose variability of this dataset is limited compared to other datasets [28, 40], and annotations are rather inaccurate as discussed in [18, 30].

We show a comparison to different state-of-the-art methods in Table 2. Again, our method shows state-of-the-art accuracy. However, the gap to other methods is much smaller. This may be attributed to the fact that the dataset is much easier, with smaller pose variations [30], and due to errors in the annotations for the evaluation [18, 30].

| Method | Average 3D error |
| --- | --- |
| Oberweger *et al*. [18] (DeepPrior) | 10.4mm |
| Deng *et al*. [2] (Hand3D) | 10.9mm |
| Tang *et al*. [34] (LRF) | 12.6mm |
| Wan *et al*. [42] | 8.2mm |
| Zhou *et al*. [48] (DeepModel) | 11.3mm |
| Sun *et al*. [28] (HPR) | 9.9mm |
| Wan *et al*. [41] (Crossing Nets) | 10.2mm |
| Fourure *et al*. [4] (JTSC) | 9.2mm |
| Krejov *et al*. [12] (CDO) | 10.5mm |
| **This work (DeepPrior++)** | **8.1mm** |

Table 2: Comparison with state-of-the-art on the ICVL dataset [34]. We report the average 3D error in mm.

In Figure 6 we compare DeepPrior++ to other methods on the ICVL dataset [34]. Our approach performs similar to the works of Guo *et al*. [6], Wan *et al*. [42], and Tang *et al*. [35], all achieving state-of-the-art accuracy on this dataset. This might be an indication that the performance on the dataset is saturating, and the remaining error is due to the annotation uncertainty. This empirical finding is similar to the discussion in [30]. Although Tang *et al*. [35] performs slightly better in some parts of the curve in Figure 6, our approach performs significantly better on the NYU dataset, as shown in Figure 5.

### 5.4. MSRA Dataset

The MSRA dataset [28] contains about 76k depth frames. It was captured using a time-of-flight camera. The dataset comprises sequences from 9 different subjects. We follow the common evaluation protocol [5, 29, 41] and perform a leave-one-out cross-validation: We train on 8 different subjects and evaluate on the remaining subject. We repeat this procedure for each subject and report the average errors over the different runs.

A comparison of the average 3D error is shown in Table 3. Again, DeepPrior++ outperforms the existing methods by a large margin of 3mm. In Figure 7, DeepPrior++ also outperforms all other methods on the plotted metric,

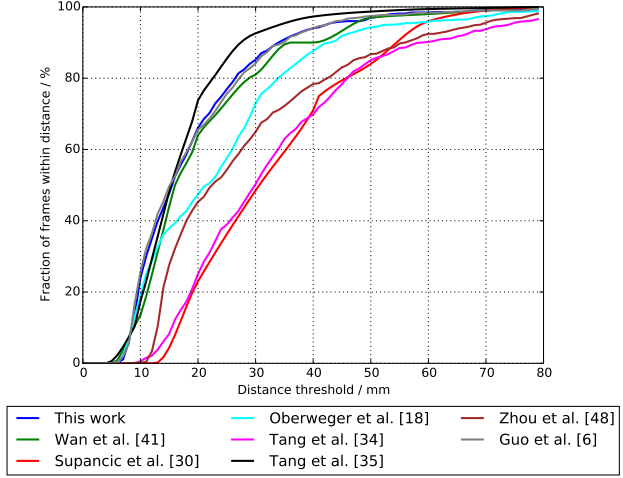

Figure 6: Comparison with state-of-the-art on the ICVL dataset [34]. We plot the fraction of frames where all joints are within a maximum distance from the ground truth. Several works show a similar error curve, which can be an indicator for saturating performance for this dataset. (Best viewed in color)

which shows that it is also able to handle different users' hands.

| Method | Average 3D error |
| --- | --- |
| Ge *et al*. [5] | 13.2mm |
| Sun *et al*. [28] (HPR) | 15.2mm |
| Wan *et al*. [41] (CrossingNets) | 12.2mm |
| Yang *et al*. [45] | 13.7mm |
| **This work (DeepPrior++)** | **9.5mm** |

Table 3: Comparison with state-of-the-art on the MSRA dataset [28]. We report the average 3D error in mm. DeepPrior++ significantly performs better than all other methods for this dataset.

### 5.5. Ablation Experiments

We performed additional experiments to show the contributions of our modifications. We evaluate the modifications on the NYU dataset [40], since it has the most accurate annotations, with diverse poses, and two different users for evaluation.

#### 5.5.1 Training Data Augmentation

In order to evaluate the contribution of the training procedure, we tested the different data augmentation schemes.

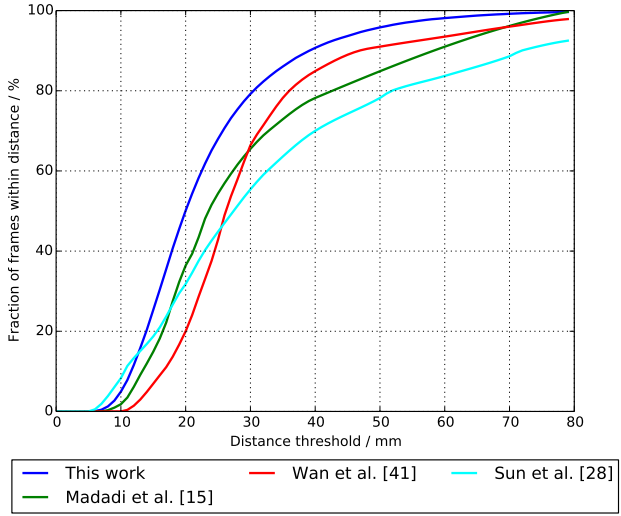

Figure 7: Comparison with state-of-the-art on the MSRA dataset [28]. We plot the fraction of frames where all joints are within a maximum distance from the ground truth. A larger area under the curve indicates better results. Our approach significantly outperforms current state-of-the-art discriminative approaches on this dataset. (Best viewed in color)

The results are shown in Table 4. Using data augmentation results in an increase in accuracy over 7mm. Most importantly, augmenting the hand translation accounts for errors in the hand detection part, and augmenting the rotation accounts for rotated hand poses, thus effectively enlarging the training poses.

Although augmenting the scale only does not help as much as augmenting translation or rotation on the NYU dataset, it can help in cases where the size of the users' hand is not accurately determined, *i.e.* a new user in a practical application. Interestingly, computing the prior from augmented the 3D hand poses is very important as well. If the data is augmented, but the prior is computed from the original 3D hand poses, the accuracy is worse compared to no data augmentation, since the prior is not expressive enough to capture the variances of the augmented hand poses.

#### 5.5.2 Hand Localization

Further, we evaluate the influence of the hand localization on the final 3D joint error. For this experiment, we use the ResNet architecture and all data augmentation. The results are shown in Table 5. The highest accuracy can be achieved using the ground truth location of the hand, which is not feasible in practice, since real detectors do not provide perfect hand localization. This indicates, that there is still room for improvement by using a more accurate 3D hand localization method.

| Augmentation | Average 3D error |
|---|---|
| No augmentation | 19.9mm |
| Translation (T) | 14.7mm |
| Rotation (R) | 13.8mm |
| Scale (S) | 17.1mm |
| All (R+T+S) | 12.3mm |
| All (R+T+S) & no prior aug. | 21.7mm |

Table 4: Effects of the new training procedure on the NYU dataset [40]. By using different data augmentation methods, the accuracy can be significantly increased. In the first row we do not use any data augmentation. In the last row we apply augmentation on the training data, but not for computing the pose prior, showing the importance of having a good pose prior.

Starting with the very simple center of mass localization and by refining the estimated center of mass localization, this step decreases the 3D localization error by almost 20mm. This in turn improves further the final average 3D pose error by over 1mm.

| Localization | Avg. 3D pose error | Loc. 3D error |
|---|---|---|
| CoM | 13.8mm | 28.1mm |
| Refined CoM | 12.3mm | 8.6mm |
| Ground truth | 10.8mm | 0.0mm |

Table 5: Impact of hand localization accuracy on NYU dataset [40]. The ground truth localization gives the lowest 3D pose error, but this localization is not applicable in practice. Our refinement of the commonly used center of mass localization (CoM) improves the accuracy by over 1mm.

#### 5.5.3 Network Architecture

We evaluate the impact of the different network architectures in Table 6. We use the refined hand localization and all data augmentation for training both networks. The improved training procedure and better localization already improve the results for the original architecture by more than 3mm (19.8mm from [18]). Using the proposed ResNet architecture, the accuracy can be improved by another 4mm on average, due to the higher capacity of the model. We also evaluated the original architecture, but changed the convolutional layers such that they use the same number of filters as the ResNet architecture, but this architecture is still inferior to the ResNet.

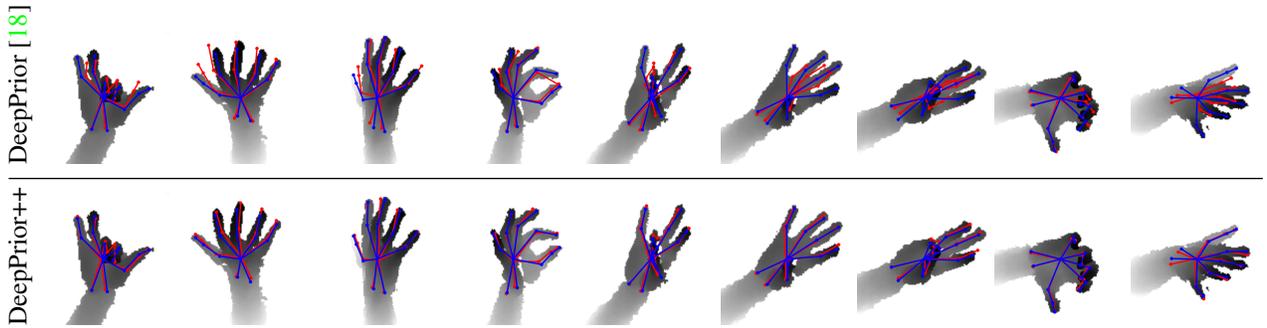

Figure 8: Qualitative comparison between DeepPrior and DeepPrior++ on the NYU dataset [40]. We show the inferred 3D joint locations projected on the depth images. Ground truth is shown in blue, the predicted poses in red. The results provided by DeepPrior++ are significantly better than the results from the original DeepPrior, especially on complex poses. (Best viewed in color)

The ResNet architecture is slower than the original implementation, however it is still able to run at over 30fps on a single GPU, making it applicable to realtime applications.

| Architecture | Average 3D error | fps |
|---|---|---|
| Original [18] | 16.6mm | 100 |
| Original with more filters | 13.7mm | 80 |
| ResNet | 12.3mm | 30 |

Table 6: Impact of network architecture on the NYU dataset [40]. The more recent ResNet architecture performs significantly better than the original network architecture, even when using the same number of filters as ResNet for the Original architecture (*Original with more filters*). Most importantly, we can still maintain realtime performance with 30fps in our hand tracking application.

### 5.6. Qualitative Evaluation

We show several qualitative results in Figure 8, where we compare to the original DeepPrior [18]. In general, DeepPrior++ provides significantly better results compared to the original DeepPrior, especially on highly articulated poses. This can be attributed to the data augmentation and better localization, but also to the more powerful CNN structure, which enables the CNN to learn highly accurate poses for complex articulations.

## 6. Discussion and Conclusion

Since the publication of DeepPrior, other works on pose estimation introduced a pose prior in a Deep Learning framework, showing the importance of such prior:

- [22] proposed to replace the linear transformation computed by the PCA by an encoder. This encoder is trained first, together with a decoder, to predict a compact representation of the pose. As the decoder has a more complex form, it brings some improvement in accuracy.

- [39] considers human pose estimation and also uses an auto-encoder, but to compute a pose embedding of *larger* dimensions than the original pose, which appears to significantly improves the accuracy in the case of body pose estimation.

- [49] learns a pose prior for estimating the 3D hand joint locations from 2D heatmaps by factorizing the prior into canonical coordinates and a relative motion, while our prior learned with PCA does not distinguish between the two.

Maybe a high-level conclusion of the work presented in this paper is that our community should be careful when comparing approaches: By paying attention to its different steps, we were able to make DeepPrior++ perform significantly better than the original DeepPrior and performs similarly or better than more recent works, while the key ideas are the same for the two methods.

**Acknowledgment:** This work was partially funded by the Christian Doppler Laboratory for Semantic 3D Computer Vision.